\documentclass[10pt,twocolumn,letterpaper]{article}

\usepackage{iccv}
\usepackage{times}
\usepackage{epsfig}
\usepackage{graphicx}
\usepackage{amsmath}
\usepackage{amssymb}

\usepackage{algorithmic}
\algsetup{linenosize=\small}
\usepackage[noend]{algorithm2e}
\SetAlgoSkip{}

\usepackage[pagebackref=true,breaklinks=true,letterpaper=true,colorlinks,bookmarks=false]{hyperref}

\iccvfinalcopy 

\def\iccvPaperID{2626} 

\begin{document}

\title{Generic Tubelet Proposals for Action Localization}

\author{Jiawei He\\
Simon Fraser University\\
{\tt\small jha203@sfu.ca}
\and
Mostafa S. Ibrahim\\
Simon Fraser University\\
{\tt\small mostafa.saad.fci@gmail.com}
\and
Zhiwei Deng\\
Simon Fraser University\\
{\tt\small zhiweid@sfu.ca}
\and
Greg Mori\\
Simon Fraser University\\
{\tt\small mori@cs.sfu.ca}
}

\maketitle
\begin{abstract}
   We develop a novel framework for action localization in videos.  We propose the Tube Proposal Network (TPN), which can generate generic, class-independent, video-level tubelet proposals in videos. 
   The generated tubelet proposals can be utilized in various video analysis tasks, including recognizing and localizing actions in videos. In particular, we integrate these generic tubelet proposals into a unified temporal deep network for action classification.  Compared with other methods, our generic tubelet proposal method is accurate, general, and is fully differentiable under a smoothL1 loss function.  We demonstrate the performance of our algorithm on the standard UCF-Sports, J-HMDB21, and UCF-101 datasets.
   Our class-independent TPN outperforms other tubelet generation methods, and our unified temporal deep network achieves state-of-the-art localization results on all three datasets.

   

\end{abstract}

\section{Introduction}

Action understanding in videos has many applications such as sports analysis, video surveillance, and content-based retrieval.  Detailed understanding  requires localizing and classifying human actions in videos, answering the questions ``what is the action in the video?" and also ``where exactly is the action in the video?"

Breakthroughs in image understanding have been achieved using the Convolutional Neural Network (CNN) structure.  The efficacy and efficiency of such structures in the action recognition area lag behind.  Strong results have been achieved for video-level analysis by adapting CNN architectures to classification tasks (e.g.\ \cite{donahue,karpathy2014large}).  However, in action-related videos actions typically occur only within a limited spatial extent inside each frame.
CNN frameworks that represent entire video frames lack focus on relevant features, are susceptible to background clutter and camera motion, and cannot perform action localization.

Finding the action region in videos can aid in developing more accurate models.
Recent advances in object detection using region proposal networks (e.g.\ \cite{fasterrcnn}) have triggered a flurry of research in action localization.
A major challenge is addressing the large search space of video content in terms of analyzing potential regions as action proposals.  Most recent work on spatial-temporal localization shares a similar framework \cite{actiontubes,actiontubeseccv,actiontubesbmvc}. 
These approaches first detect specific actions at a frame level by utilizing 2-D detection networks (R-CNN \cite{RCNN} or faster R-CNN \cite{fasterrcnn}). The class-specific detections in each frame are then linked or tracked, resulting in class-specific action tubes.
To summarize, these methods tend to treat a video as a set of independent images. Therefore, action localization is performed on each image separately, losing much potential to exploit temporal relationships in each video.

\begin{figure}[!t]
\centering
\includegraphics[width=0.5\textwidth]{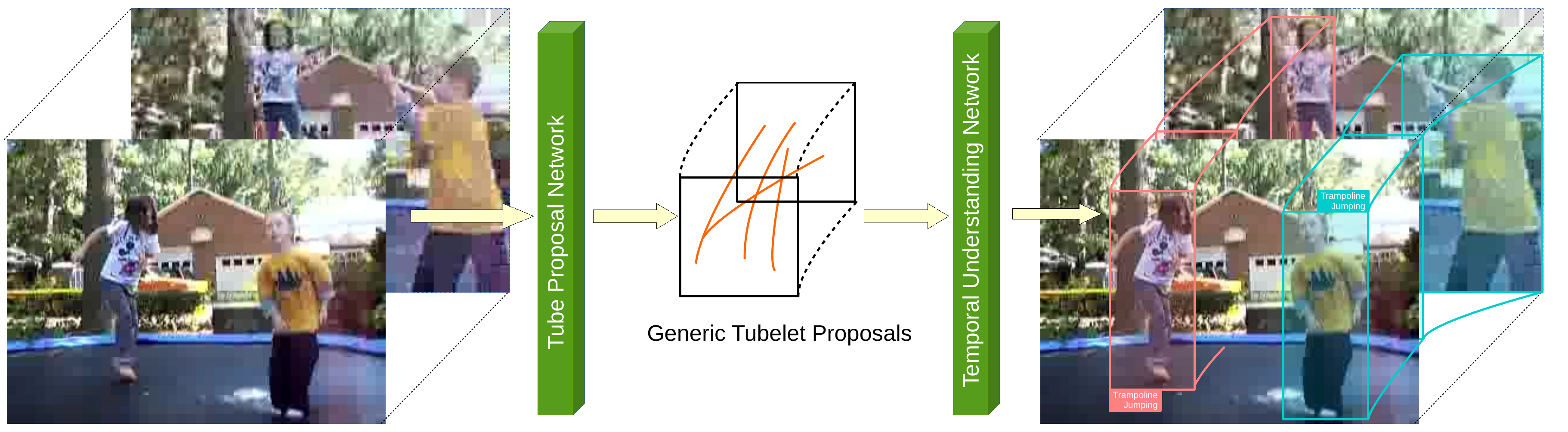}
\caption{Action localization with our Tube Proposal Network produces generic class-independent tubelets. The tubelets are classified with a Temporal Understanding Network that can perform detailed spatio-temporal analysis. } \label{fig:overview}
\end{figure}

\begin{figure*}[ht]
\centering
\includegraphics[width=\textwidth]{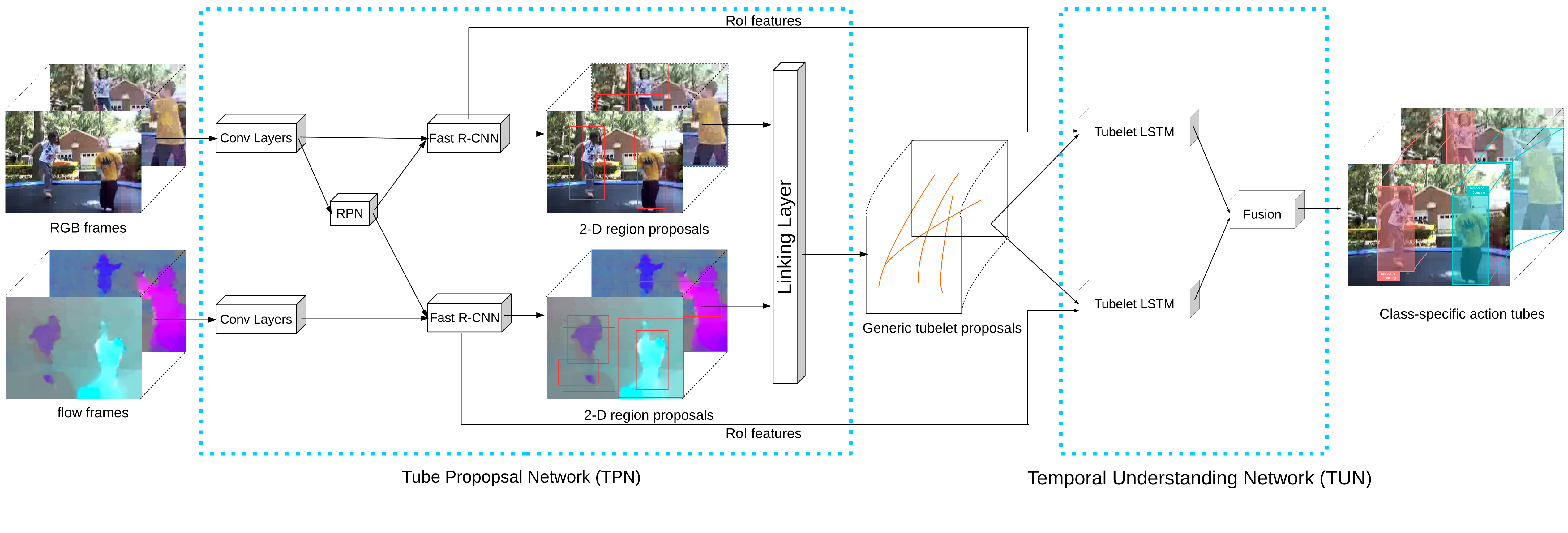}
\caption{The proposed framework consists of two components: a tubelet proposal network (TPN) and a temporal understanding network (TUN). A TUN based on fused LSTMs is used to classify the generic class-independent tubelet proposals generated by the TPN.}\label{fig:pipeline}
\end{figure*}



Based on the discussion above, we propose a novel framework that can utilize both spatial and temporal information to spatially localize and classify actions in a video sequence.
The framework can automatically generate multiple generic class-independent tubelet proposals in each video.  This approach with generic tubelets alleviates important shortcomings with previous methods.  Notably, the challenging classification step can be conducted using all available temporal information in a tubelet. This obviates the need to make difficult classification decisions at a frame level, allows correction for camera motion, and discards background clutter.  Further, a single set of tubelets is generated in one pass, usable for all further analysis.  Finally, the proposal method is generic and can be easily integrated into other spatio-temporal deep networks.

We term our framework as the combination of a \textit{ Tube Proposal Network (TPN)} and a \textit{Temporal Understanding Network (TUN)}. 
The TPN generates multiple video-level, generic, class-independent tubelet proposals. A tubelet proposal is  a spatio-temporal region proposal, analogous to the generic region proposals used in object detection.
These generic tubelet proposals are then fed into the TUN, which can incorporate temporal information for action analysis. A variety of temporal models such as temporal pooling~\cite{karpathy2014large}, LSTM~\cite{hochreiter1997long}, or C3D~\cite{Tran15} could be easily used in the TUN. In this work, a temporal model based on fused LSTMs is utilized as the TUN to classify the generated generic tubelet proposals.

In summary, the main contributions of this paper are: (1) a novel tubelet proposal network generating generic video-level tubelet proposals to enable person-centric action recognition; (2) a general TPN / TUN structure that can be easily instantiated using a variety of video understanding network architectures; (3) development of a specific TUN that can exploit spatio-temporal features for classification in each tubelet;  and (4) state-of-the-art performance on the three standard datasets for spatial action localization.  


	




\section{Related Work}
\label{sec:rel_work}

Inspired by the promising performance of CNNs in image classification and object detection, deep learning architectures have been increasingly utilized in video analysis.  We start this section by reviewing relevant work in general action recognition and object detection, before turning our attention to action localization.

\noindent \textbf{Action Recognition:}
Extensive research effort has been devoted to action recognition. Recent surveys~\cite{Aggarwal11Review,Weinland11Review} summarize methods based on hand-crafted features.

Following the impressive performance of CNNs in object recognition, deep learning approaches were applied to action recognition. Karpathy et al.~\cite{karpathy2014large} use frame stacks as input to a network to learn a deep temporal representation. 
Simonyan and Zisserman~\cite{Simonyan14NIPS} train two separate ConvNets (one for an appearance feature using RGB images and another for motion features using optical flow images), and fuse their results to achieve performance competitive with the Improved Dense Trajectory approach~\cite{Wang13IDenseTraj}. Later work builds on this two-stream structure to boost accuracy by utilizing motion cues. Zha et al.~\cite{Zha15VidClassification} aggregate CNN features from VGGnet~\cite{Simonyan14VGG} with dense trajectory features~\cite{Wang13IDenseTraj} using simple pooling strategies to produce spatio-temporal features for video classification. An in-depth exploration of the right choices for the classification pipeline (pooling, feature  normalization, layers, classifiers) led to significant performance gain compared to contemporary approaches.

\noindent \textbf{Object Detection:}
The object detection task is typically posed as generating a bounding box around each object instance in an image. 
Hand-crafted features such as SIFT or HOG were utilized in the pre-AlexNet era.  The R-CNN approach~\cite{RCNN} was the first to successfully deploy CNNs for object detection.  It extracts a set of region proposals from an image using selective search~\cite{van11SelectiveSearch}, which are then fed through a fine-tuned CNN network separately to classify the object inside the proposed region.
Although the approach achieved impressive performance, it is slow during both training and testing. 
Two critical improvements followed that approach: Fast R-CNN~\cite{GirshickICCV14FastRcnn}, where the features of the region proposals are computed efficiently using a shared network per image; and Faster R-CNN~\cite{fasterrcnn}, where a region proposal network (RPN) is introduced to efficiently generate the candidate regions.
 
\noindent \textbf{Action Localization:}
The action localization task is similar to object detection -- the goal is to spatially/temporally localize a recognized action within a video, often using a per-frame bounding box representation.

Seminal work includes Ke et al.~\cite{Ke07EventDetection}, who proposed a template-based method to build models for human action localization in crowded areas.  These hand-labeled templates are matched based on shape and motion features against over-segmented spatio-temporal clips. Shechtman and Irani~\cite{Shechtman05EventDetection} proposed a space-time correlation method for actions in video segments with an action template based on enforced consistency constraints on the local intensity patterns of spatio-temporal tubes.  
Lan et al.~\cite{lan} used latent SVM learning to jointly detect and recognize actions in videos based on a figure-centric visual word representation. Van Gemert et al.~\cite{densetraj} utilize dense trajectory features for region proposal and classification.

A promising direction is recent attempts to learn deep networks for action localization.  Typically, 2-D action regions are detected in each frame and linked to generate 3-D action volumes. One of the leading directions in this vein is Gkioxari and Malik's work~\cite{actiontubes} that generates action tubes based on appearance and motion cues from two-stream CNN networks. 
In their approach, candidate regions are generated from each frame (using selective search~\cite{van11SelectiveSearch}) and only the regions with enough motion saliency are kept. These regions are classified and scored using an SVM and linked in time to build the final action tubes.

Several approaches~\cite{weinzaepfel,actiontubesbmvc,actiontubeseccv} present interesting improvements on this direction. Weinzaepfel et al.~\cite{weinzaepfel} uses a linking approach based on tracking, with region candidates generated using EdgeBoxes~\cite{ZitnickECCV14edgeBoxes}.
Saha et al.~\cite{actiontubesbmvc} proposes a better method to fuse the two stream detections in each frame and smooth path labeling.
Peng and Schmid~\cite{actiontubeseccv} further expands the two-stream structure into four streams by dividing the proposal region in each frame into upper and lower regions.
These two methods improve performance significantly compared with \cite{actiontubes}, by updating the CNN from Alexnet (used in \cite{actiontubes}) to VGGnet, and replacing selective search with the region proposal network for more accurate 2-D region proposal generation.  Further improvements are possible by class-specific approaches, e.g. \cite{actiontubesbmvc} obtains a $6\%$ increase in performance via a class-dependent hyper-parameter. 

Although the recent state-of-the-art methods \cite{actiontubes,actiontubesbmvc,actiontubeseccv} are very effective, they all treat each frame independently in many classification stages, reducing the temporal information present in a video sequence.  Different from these methods, we first generate a set of generic class-independent tubelet proposals for each video, which are then classified using a temporal model. Our model thereby can build detailed person-centric models for exploiting both spatial and temporal information for localizing and classifying actions.

\section{Action Localization via Generic Tubelets}
\label{sec:method}

A key challenge in action recognition is that individual frames are ambiguous -- given a single image of a person in an upright pose, this could correspond to a person walking, standing, kicking a ball, or myriad other actions that include a fleeting moment of similarly neutral posture.  Furthermore, human action involves a variety of different poses.  Accurately detecting people contorted into varied poses is a challenge, and utilizing all available ``human action-like" data to build a detector seems favourable to building action-specific human detectors.

For these reasons, we advocate the building of generic tubelets for action localization.  Generic tubelets can leverage all available human action training data for their detection.  Person-centric tubelets can be constructed, then classified into action categories in their entirety.  In order to accurately recognize actions in video, it makes sense to delay making hard classification decisions until all information present in video sequence has been observed.

We operationalize these ideas via our proposed  approach, consisting of a \textit{tube proposal network} (TPN) and a \textit{temporal understanding network} (TUN), as illustrated in Fig.~\ref{fig:pipeline}. The TPN generates a set of  video-level generic tubelet proposals and the TUN utilizes these tubelet proposals to perform more sophisticated temporal understanding in each video. 
The output of the entire framework is a set of labeled action tubes for each video. 
The proposed framework is described in detail below.

\subsection{Tube Proposal Network}

The goal of the TPN is to generate spatio-temporal region proposals (tubelet proposals) where it is likely that an action occurred. Importantly, at this stage of analysis we are not concerned with the action category, just whether {\it any} action is taking place at this location.

A tubelet proposal is constructed by linking frame-level spatial proposals in time. Frame-level proposals are represented as bounding boxes, ideally covering the action region. The obvious advantage of this method is that the tubelet proposal moves with the person in time, focusing on the action region specifically. This mechanism enables person-centric region-of-interest (RoI) feature extraction, which we show leads to improved classification accuracy. Furthermore, since the proposals we link are generic, the localization accuracy can be improved (see Sec.\ \ref{sec:RPG}). As a result, our framework improves both localization and classification accuracy.
The TPN consists of two parts: 2-D generic region proposal generation, and 3-D generic tubelet proposal generation.


\subsubsection{2-D Region Proposal Generation}
\label{sec:RPG}

\begin{figure}[t]
\centering
\includegraphics[width=0.45\textwidth]{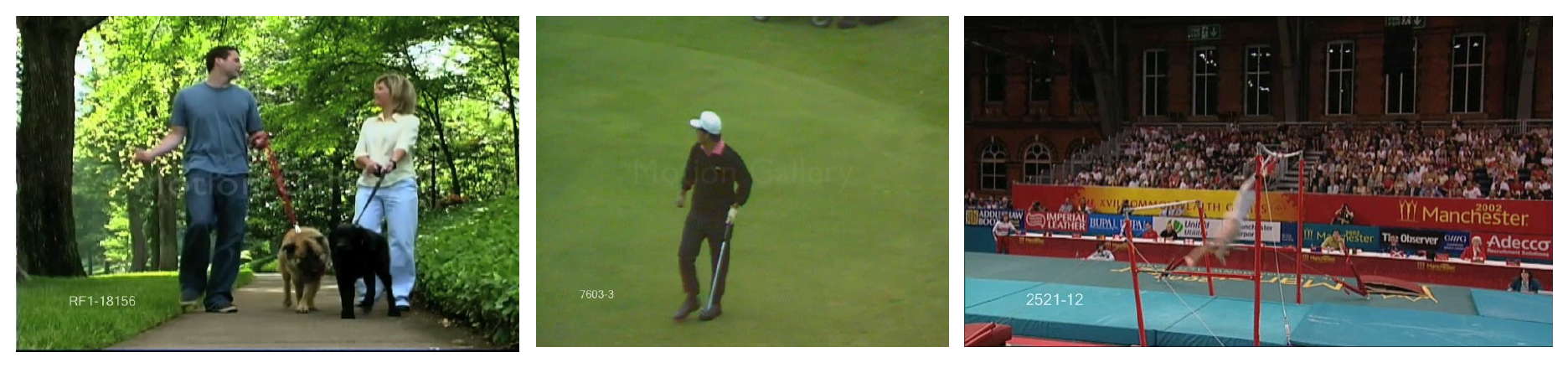}
\caption{Examples of hard  cases for class-dependent action proposals. The first image action label is \textit{walking}, while the second one is \textit{running}. The action region in the third image has significant motion blur. The three frames come from the UCF-Sports dataset.} \label{fig:hardcase}
\end{figure}

To generate accurate video-level tubelet proposals, we first acquire a set of 2-D action region proposals in each frame, which breaks down the task to frame-level action detection. A typical object detection framework in images consists of two parts: region proposal generation / sliding window, and a network to classify and regress the proposals into final bounding boxes with labels. The regressor in recent object detection works is class-dependent: each region proposal will be regressed into a set of bounding boxes, based on classification. Therefore, it is crucial to classify the region correctly. However, this is not practical in action analysis tasks since an action often occurs over a sequence of frames. When examining a single frame, it can be difficult to discern the action category, as per the examples in Fig.~\ref{fig:hardcase}.  As a result, a class-specific regressor will face challenges in creating accurate spatial proposals that are consistent across time and action category. 

However, a generic, class-agnostic proposal mechanism can alleviate these shortcomings.  As long as a bounding box appears to contain human action, it should be included as a potential region for {\it some} action to have occurred.  The actual decision on which action it is can be deferred to a future analysis step.

As an example, consider the second image in Fig.~\ref{fig:hardcase}, which could reasonably be proposed as either a \textit{walking}, \textit{standing}, or \textit{running} region. Based on this intuition, we perform a generic class-agnostic region proposal and regression, by treating the regressed bounding box with the highest action score for each proposal as our generic region proposal. The aforementioned discussion shows that treating the regressed bounding boxes as generic region proposals is essential and effective.

For implementation, we utilize the faster R-CNN~\cite{fasterrcnn} framework as the 2-D region proposal generator. Note that although the class-specific action scores are not used to generate tubelet proposals, they are saved to help with classification, as described in Sec.\ \ref{sec:TUN}.



\begin{table*}[t]\tiny
\caption{Ablation study of our proposed framework on J-HMDB21 (split 1).}
\label{tab:ablation}
\setlength{\tabcolsep}{3.6pt}
\begin{center}
 \begin{tabular}{|c| c c c c c c c c c c c c c c c c c c c c c  |c|} 
 \hline
video-AP & brush hair & catch & clap & climb stairs & golf & jump& kick ball&	pick&	pour& 	pullup&	push&	run	&shoot ball&	shoot bow&	shoot gun&	sit& 	stand&	swing baseball&	throw&	walk&	wave & mAP\\ [0.5ex] 
 \hline\hline
 RGB  w/o & 79.95 & 30.60 & 65.79 & 70.51 &92.78	&8.15&	38.03&	75.02&	95.72&	100.0&	87.81&	35.72&	34.16&	100.0	&72.67&	24.72&	27.95&	68.42&	22.50&	50.86&	69.80& 59.58 \\ 
  \hline
 
 Flow  w/o & 87.89	&47.20	&86.08	&61.06&	100.0	&59.22&	27.57	&75.35&	88.29&	98.13	&82.14	&45.52	&39.67	&96.24	&64.33	&86.61	&87.99	&94.17	&28.80	&77.45&	18.43& 69.15 \\ 
 \hline
 
 Fuse  w/o & 92.83&	50.75&	88.86&	71.83&	100.0&	41.79&	40.25	&81.41&	93.94&	99.63&	90.85&	55.25&	46.21&	100.0&	74.27&	71.35&	83.70&	94.77&	31.81&	77.63&	48.99& 73.15 \\ 
 \hline
 \hline
 
 RGB with TUN &  89.10	&24.48&	86.84&	70.37	&94.10&	5.60	&95.45&	93.56&	100.0	&100.0	&95.33	&74.10	&29.65	&100.0&	71.13	&28.57&	26.50&	79.15&	22.43&	39.07&	63.64 &66.15\\ 
 \hline

 Flow with TUN & 99.36	&47.10&	89.87&	58.62&	100.0&	59.32&	25.08&	100.0&	93.16&	99.63	&98.81&	40.35&	46.26&	94.62	&68.17&	97.22&	88.27&	92.29&	28.69&	85.16&	11.94& 72.57\\
   \hline
 
 Fuse with TUN & 100.0&	46.09&	97.09&	69.93&	100.0&	29.33&	91.42&	100.0&	100.0&	100.0&	100.0&	78.36&	40.72&	100.0&	77.10&	54.49&	75.91&	93.79&	29.35&	89.27&	45.88 & \textbf{77.08}\\
 \hline

\end{tabular}
\end{center}
\end{table*}

\subsubsection{Video-level Generic Tubelet Proposal Generation}
\label{sec:VGT}

Given the 2-D region proposals, the natural next step would be to link them into tubelets. We build our tubelets in an incremental fashion by linking existing tubelets to 2-D region proposals in the subsequent frame.  Since there are multiple region proposals in each frame, it is important to identify which region proposals should be linked given a tubelet proposal from previous time steps. The goal is to ensure that a tubelet does not shift between different people over time.


In this work, we utilize a simple yet effective greedy linking algorithm to generate tubelet proposals.
Firstly, the {\it objectness} score $o^{i}_{t}$ of a 2-D region proposal $i$ in frame $t$ is defined as $o^{i}_{t} = \underset{c \in C}{max}(s^{i}_{t}(c))$, where $s^{i}_{t}(c)$ denotes the \textit{class-dependent} action score of region $i$ in frame $t$, for class $c \in C$, the set of action classes in the data set. With this per-region objectness score, we describe the linking algorithm as follows. Consider two consecutive frames at time $t-1$ and $t$, and assume $R_{t}^{i}$ is spatial location of the $i^{th}$ region proposal in frame $t$. 
The linking score $\ell (R_{t}^{i},R_{t-1}^{j})$ between two regions $R_{t}^{i}$ and $R_{t-1}^{j}$ is defined to be
\begin{equation}\label{eq:ls}
\ell (R_{t}^{i},R_{t-1}^{j}) = o_{t}^{i} + o_{t-1}^{j} + \frac{\cap}{\cup}(R_{t}^{i},R_{t-1}^{j})
\end{equation}
where $\frac{\cap}{\cup}$ denotes the intersection over union of the two regions.  At time-step $t$, for each tubelet proposal, linking scores are computed between the region in that tubelet at frame $t-1$ and all possible proposals in the frame $t$. The proposal with largest $\ell (\cdot,\cdot)$ is linked to the tubelet.
For efficiency, we greedily prune to the top-$K$ ($K=10$ in experiments) tubelet proposals at each time-step $t$, starting from time-step $2$, based on the tubelet score $\tau_{t}^{m}$ for each tubelet $m$:
\begin{equation} \label{eq:obj}
\tau_{t}^{m} = \sum_{k=2}^{t} \ell (R_{k}^{i(m,k)},R_{k-1}^{i(m,k-1)})
\end{equation}
where $i(m,k)$ indexes the regions $R_{k}^{i(m,k)}$ for tubelet $m$ at each time step $k$.
A final video-level non-maximum suppression is run to produce a set of class-independent tubelet proposals for the video. Algorithm~\ref{alg:tubelets} summarizes this process.

\begin{algorithm}[]

\SetAlgoLined
\KwIn{Proposals $R_{t}^{i}$ in each frame $t$ }

 \For{$t = 1 \ldots T$}{
  \If{t==1}{
   \For{each $R_{t}^{i}$}{
    init new tubelet from $R_{t}^{i}$\;}}
   \If{$t > 1$}{
    { \For{each tubelet $m$ in frame $t-1$}{
    Link i with best $R_{t}^{j}$ by Eq. \ref{eq:ls} \; }
    }
    keep top-K tubelets as scored by Eq. \ref{eq:obj} \;}
 }
 
 video-level NMS\;
 
\KwOut{tubelet proposals}

 \caption{Tubelet Generation Algorithm}
 \label{alg:tubelets}
\end{algorithm}

\subsection{Temporal Understanding Network}
\label{sec:TUN}

Videos contain rich information in both spatial and temporal dimensions. Simple temporal approaches, such as averaging frame-level spatial features, are not sufficient to represent temporal information. Therefore, we input the video-level tubelet proposals to a temporal understanding network (TUN) to perform more sophisticated temporal analysis. Various temporal network structures can be easily utilized here, e.g., temporal pooling, LSTM, C3D, etc. In this work, we build an LSTM-based structure to classify the generic tubelets.




In order to utilize all available information to facilitate action classification, we fuse the frame-level action score ${s}^{i(m,t)}_{t}(c)$ from the TPN  with the one $\hat{s}^{i(m,t)}_{t}(c)$ from the LSTM  for the same tube $m$ to obtain video-level information. Therefore, the class-specific action score $S_{tube}^{m}(c)$ for a tubelet $m$ is calculated by averaging the scores of TPN and LSTM from all frames in the same tubelet:




\begingroup\makeatletter\def\f@size{9}\check@mathfonts
\begin{equation} \label{eq:lamdba1}
S_{tube}^{m}(c) = \frac{\lambda_{1}}{T} \sum_{t=1}^{T} s_{t}^{i(m,t)}(c)
     +   \frac{1-\lambda_{1}}{T} \sum_{t=1}^{T} \hat{s}_{t}^{i(m,t)}(c) 
\end{equation}  
The parameter $\lambda_{1}$ is  set to $2/3$ in this work, empirically.

Since the LSTM and the TPN share the convolution layers and fully connected layers (as shown in Fig.~\ref{fig:pipeline}), the region-specific feature, namely, the fc7 feature vectors of selected region $R^{i(m,t)}_{t}$ of tubelet $m$ in TPN is given into the LSTM as input. 
Region-specific feature extraction is proved to be effective for classification in Sec. \ref{sec:j-hmdb}

\subsection{Two-stream Network Structure}
\label{sec:fusion}


We further present a generalization of our approach to a two-stream structure that utilizes both RGB frames and optical flow input.  Optical flow often proves to be effective in action classification since it uses motion information, complementary to the appearance cues in the RGB stream. 

The main advantage for optical flow is in more accurate action classification\footnote{Empirically, we noticed that a RPN trained on flow images is not effective. The simple reason is that optical flow only focuses on moving regions, and people performing actions are occasionally stationary. Therefore, in our work the RPN is always based on the RGB stream, and used in both the RGB and flow streams.}.  Hence, we develop a method for matching optical flow-stream regions with their counterparts obtained from the RGB stream.  The final fused score for a region will be the combination of both the RGB stream score and the optical flow stream score.
 In this work, the optical flow images are calculated following the method in \cite{opticalflow}.
There are two fusions in the network: the first one is the fusion of the 2-D region proposals in the TPN, and the second one is the fusion of the two LSTM outputs for the TUN.


\noindent \textbf{TPN Fusion:}
Although both streams share the same RPN (shown in Fig. \ref{fig:pipeline}), the regressor and classifier of each stream is learned independently. Therefore, it is important to decide
which flow-stream region proposal $\bar{R}^{j(i,t)}_{t}$ should be corresponding to the rgb-stream proposal $R^{i}_{t}$ in frame $t$. We calculate the IoU of each flow-stream proposal $\bar{R}^{k}_{t}$ ($k \in \{ 1, \ldots,K\}$, the set of proposals in flow stream) with a given rgb-stream proposal $R^{i}_{t}$ in the same frame.  The correspondence is established based on  Eq. \ref{eq:flow_fuse}:
\begin{equation} \label{eq:flow_fuse}
j(i,t) = \underset{k \in \{ 1,\ldots,K\}}{argmax} \frac{\cap}{\cup}(R_{t}^{i}, \bar{R}_{t}^{k})
\end{equation}
Now, each rgb-stream proposal has a corresponding flow-stream proposal. 
The class-specific action scores are updated by averaging over the rgb-stream proposal $i$ action score $s_{t}^{i}(c)$ with the corresponding flow-stream proposal action score $\bar{s}_{t}^{j(i,t)}(c)$ in frame $t$:
\begin{equation} \label{eq:lambda2}
s_{t}^{i}(c) = \lambda_{2} \times s_{t}^{i}(c) + (1 - \lambda_{2}) \times \bar{s}_{t}^{j(i,t)}(c)
\end{equation}
Since flow stream is more effective in classification, following \cite{donahue},  $\lambda_{2}$ is set to $1/3$.
Once a tubelet proposal is generated in the rgb stream, a corresponding flow-stream tubelet proposal is also generated based on the per-proposal correspondence in Eq. \ref{eq:flow_fuse}. 

\noindent \textbf{TUN Fusion:}
A LSTM-based TUN is performed on each stream separately, and the class-specific action scores (output of LSTM) are averaged over two streams using Eq.\ \ref{eq:lambda2}. In conclusion, the final class-specific action score of a tube contains four components: TPN and TUN action scores in both the rgb and optical flow streams.
A thorough ablation study is given in Sec.~\ref{sec:exp} examining these design choices.


\begin{table*}
\begin{center}

        \label{multiprogram}
        \begin{tabular}{|c|c|c|c|c|c|c|c|c|}
            \cline{1-9}
             & \multicolumn{2}{|c|}{UCF-Sports} & \multicolumn{5}{|c|}{J-HMDB21} & \multicolumn{1}{|c|}{UCF-101}  \\
            \cline{1-9}
             IoU threshold $\delta$ & 0.2 & 0.5 & 0.1 & 0.2 & 0.3 & 0.4 & 0.5 & 0.2 \\
            \hline
            ActionTubes \cite{actiontubes} & - & 75.8 & - & - & -&-& 53.5&-\\
            STMH \cite{weinzaepfel} & - & 90.5 & -& 63.1 & 63.5 & 62.2&60.7 & -\\
            STAT \cite{actiontubesbmvc}* & - & - & 72.65&72.63&72.59&72.24&71.50& -\\
            ST \cite{singh2016online}* & - & - & -&69.36&-&-&66.76& -\\            
            TS R-CNN\cite{actiontubeseccv}* & 94.8 & 94.8 & - & 71.1 & - &-& 70.60 & 69.78\\ 
            MR-TS R-CNN\cite{actiontubeseccv} & 94.8 & 94.7 & - & 74.30 & - &-& 73.09 & 69.56\\ 
            Our TPN* &95.8 & 95.5 & 74.68 & 74.58 & 74.08 & 73.38 & 71.99& 70.34\\
            Our TPN with LSTM & \textbf{96.0} & \textbf{95.7} & \textbf{79.84} &\textbf{79.70} &\textbf{79.28} &\textbf{78.53} &\textbf{76.96}& \textbf{71.69}\\
            \hline
            
        \end{tabular}
         \caption{Comparison to state-of-the-art methods on three datasets. Methods with * are considered as comparable basic models with similar network setup. The results on the J-HMDB21 data set are averages over all three splits.The experiments on the  UCF-101 data set are performed without temporal localization. Only \cite{actiontubeseccv} reports mAP  both with and without temporal localization, therefore we only compare with \cite{actiontubeseccv} (without temporal localization) on this dataset.}
         \label{tab:compare}
\end{center}
\end{table*}

\begin{table}[t]
\setlength{\tabcolsep}{3.6pt}
\begin{center}
 \begin{tabular}{|c| c c c c c c|} 
 \hline
 Method &  \cite{wangdense} & \cite{weinzaepfel} & \cite{actiontubes} & \cite{actiontubesbmvc} & \cite{actiontubeseccv} & Ours\\ [0.5ex]
 \hline
 Accuracy(\%) & 56.6 & 61 & 62.5 & 70.0 & 71.08 & \textbf{72.21}\\
 \hline
\end{tabular}
\caption{Classification accuracy on J-HMDB21 dataset (averaged over all three splits).}
\label{tab:jhmdb_class}
\end{center}
\end{table}

\subsection{Training}


In our experiments, the TPN and TUN are trained independently in a two-stage fashion, which introduces practical benefits in terms of reduced memory consumption, the ability to use (memory intensive) state-of-the-art network architectures, and initializing from pre-trained models.

The implementation uses the Caffe toolbox. The TPN is based on a VGG-16 pre-trained on ImageNet.  Note that the memory used to train the TPN alone is almost 11 GB (a 12 GB Titan X GPU is used in experiments).  We fine-tune the TPN on each dataset respectively. The learning rate is set to $10^{-3}$ and decreases by $0.1$ every $10k$ iterations. Training is stopped after $70k$ iterations.  Softmax loss for classification and SmoothL1 loss for regression are used.
RoI-fc7 features from VGG are used as the input of the LSTM, which has 3000 hidden nodes. The LSTM learning rate is set to $10^{-4}$, decreased by 0.1 per 10k iterations. Momentum and weight decay are set to $0.9$ and $5e^{-4}$. Softmax loss is used in training the LSTM for classifying tubelets.

Based on this training setting, the whole framework is fully differentiable under the per-frame SmoothL1 loss for regression. Therefore, the TPN and TUN could be potentially jointly optimized, trained in an end-to-end fashion.  However, as mentioned above, memory considerations preclude doing so at this time.



\section{Experimental Results and Discussion}
\label{sec:exp}

We evaluate our approach on three standard datasets: UCF Sports, J-HMDB21, and UCF-101.
On these data sets, we compare against other methods and show substantial improvement over state-of-the-art approaches.

\noindent \textbf{Datasets}. The UCF-Sports dataset consists of 150 videos with 10 different actions. 
We use the same evaluation protocol as~\cite{lan}.
The J-HMDB21 dataset contains about 900 videos of 21 different actions.
The human figure in each frame has body joint annotations, which are used to generate the bounding box for each person. A subset of 24 classes out of 101 (‘split 1’) comes with spatio-temporal localisation annotation from  \cite{singh2016online}.  We use these for measuring action localization, following the same protocol as ~\cite{singh2016online, actiontubesbmvc, actiontubeseccv}.


\noindent \textbf{Evaluation Metric}. We use \textit{video mean average precision (mAP)} as the main evaluation metric. Similar to object detection, an action tube is considered as a true positive if 1) the IoU of the  action tube and the ground-truth tube is above a threshold $\delta$, and 2) the action in the tube is correctly classified. This is the same evaluation metric as in previous action localization works such as \cite{actiontubes,actiontubesbmvc,actiontubeseccv,singh2016online}.

\subsection{Ablation Study}
\label{sec:ablation}

We conducted an ablation study to assess the utility of components of our approach.  We compare the following variants on the first split of the J-HMDB21 dataset.  Results are shown in Table~\ref{tab:ablation}.

\noindent{\textbf{RGB-stream model without TUN}}. This simple baseline averages the class-specific action scores from the fast R-CNN.
    
\noindent{\textbf{RGB-stream model with TUN}}. Generic tubelet proposals from the TPN are fed to an LSTM to improve the classification in each video.
    
\noindent{\textbf{Flow-stream model without TUN}}.  This model is similar to the RGB-stream model without TUN, with the input being the optical flow images.
    
\noindent{\textbf{Flow-stream model with TUN}}. This model is similar to the RGB-stream model with TUN, with the input being the optical flow images.
    
\noindent{\textbf{Two-stream fusion model without TUN}}. This model fuses the 2-D region proposals to generate more accurate generic class-independent tubelet proposals. 
The fused score is used to classify these tubes.
    
\noindent{\textbf{Two-stream fusion model with TUN}}. This  is our full proposed model.

\begin{figure*}[t]
\centering
\includegraphics[width=1.0\textwidth]{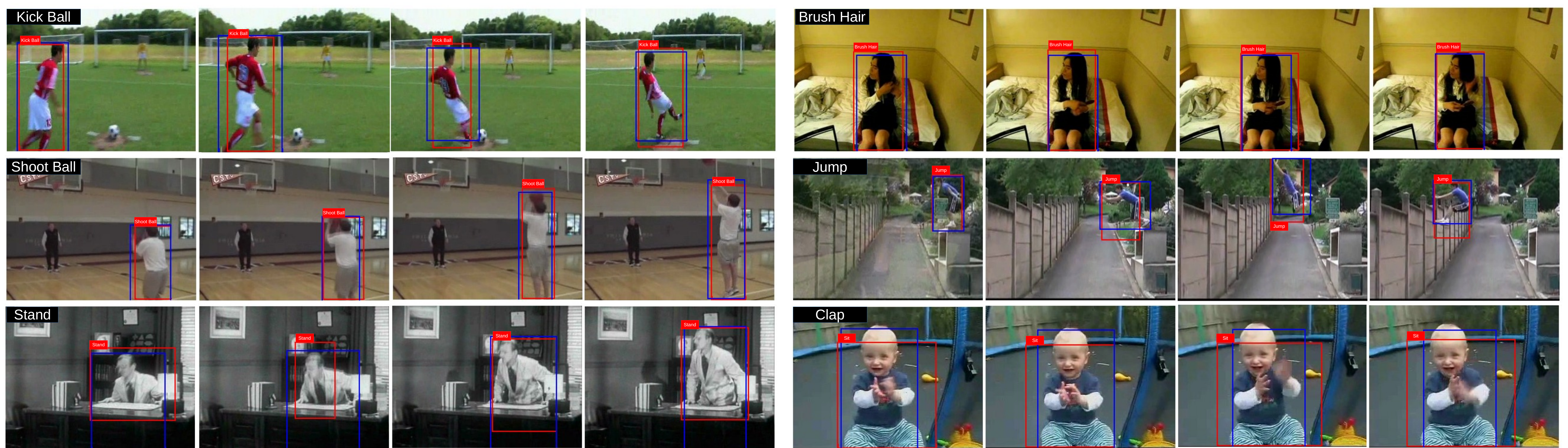}
\caption{Examples from the J-HMDB dataset. Each video is represented by four frames. The red bounding boxes are the detected regions, and the blue ones are the ground-truth regions. The first three rows show successful cases (action classification is correct and averaged localization overlap is larger than 0.5).
The bottom left one is a failed case where the averaged overlap ratio is smaller than 0.5. The bottom right one is a failed case in which the ground-truth label is \textit{Clap}, but is classified as \textit{Sit}.} \label{fig:visualization}
\end{figure*}

\begin{figure*}[t]
\centering
\includegraphics[width=1.0\textwidth]{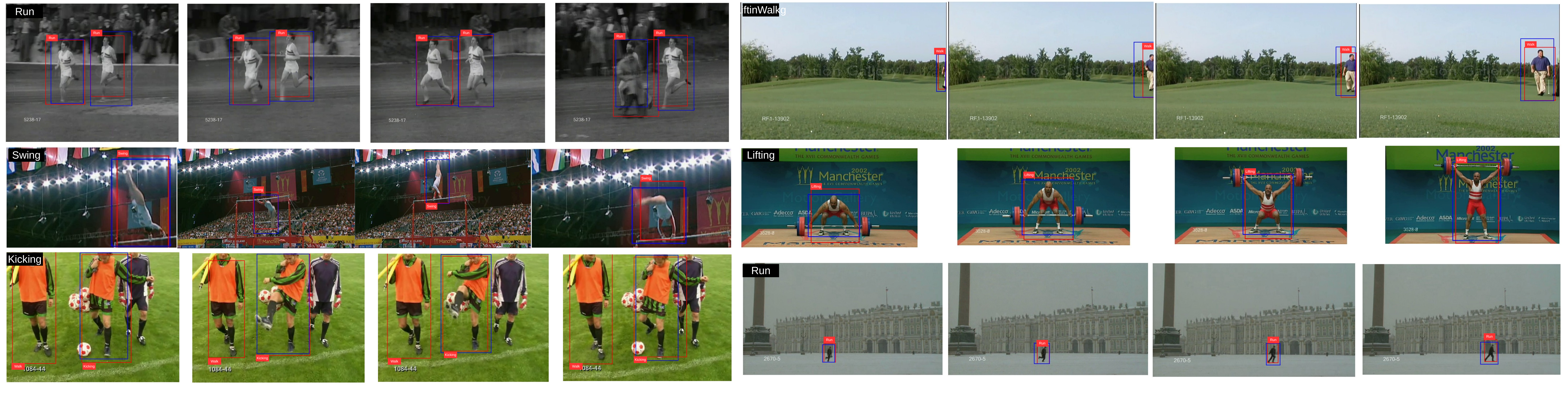}
\caption{Example action tubes from the UCF-Sports dataset. Each video is represented by four frames. The red bounding boxes are the detected regions, and the blue ones are the ground-truth regions.
As can be seen, our generated tubes are very accurate. 
Also, our framework can handle multiple-person scenarios (upper left video), fast moving actions like \textit{Swing},  and  actions with large intra-class shape variation like \textit{Lifting}.
Left bottom shows a failure case. The ground-truth of this video only focused on the middle person with label \textit{Kicking}. Our algorithm generates two high-scoring tubes for this video, one overlaps with the ground-truth region, another is the left person with label \textit{Walk}, which we think is also reasonable.} \label{fig:visualization_sports}
\end{figure*}

\begin{figure*}[t]
\centering
\includegraphics[width=1.0\textwidth]{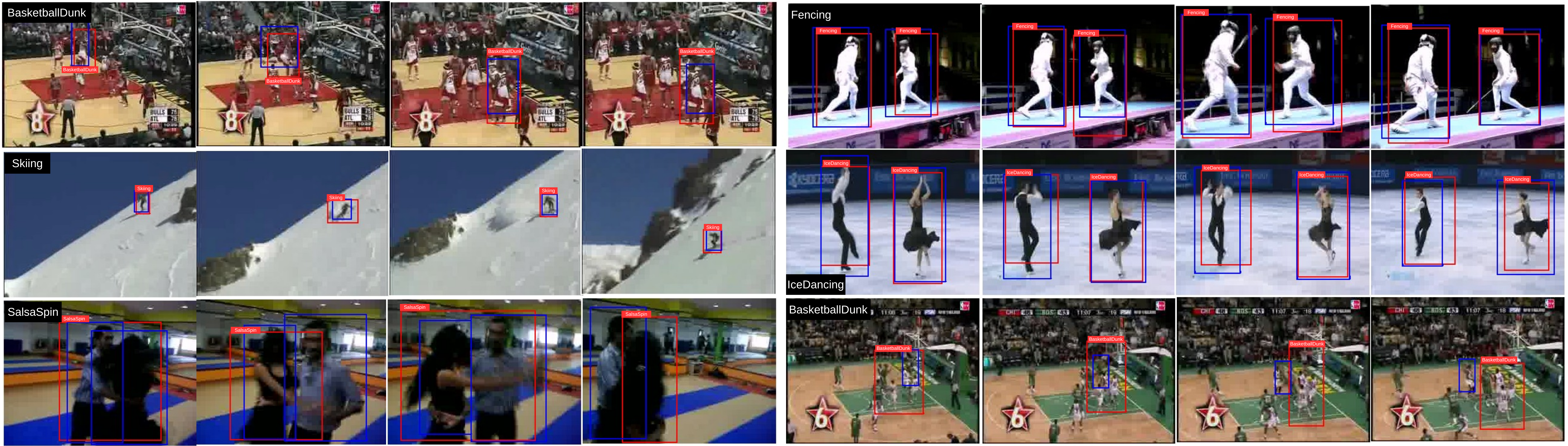}
\caption{Example action tubes from the UCF-101 dataset. Each video is represented by four frames. The red bounding boxes are the detected regions, and the blue ones are the ground-truth regions.
The first two rows show successful cases, while the last row shows failure cases. Our framework generates accurate action tubes for complicated actions. However, the bottom-left case shows that our framework tends to generate only one action tube if multiple action instances are close to each other. The bottom-right and top-left cases show performance in complicated scenes such as basketball games (top-left correct, bottom-right inaccurate localization).} \label{fig:visualization_101}
\end{figure*}

It is clear from the ablation study that each part of our model improves the tubelet generation performance. 
Without using a TUN, the fusion model alone outperforms other comparable methods.
The three methods that are considered as comparable share the same  VGG16 network structure, and all use faster R-CNN to generate 2-D region proposals/detections. In Table~\ref{tab:ablation}, the three comparable methods are marked with *.
Furthermore, adding LSTM as an example of the TUN component increases the mAP, which shows that the generated generic tubelet proposals preserve the information for the action to be correctly classified.

\subsection{Comparison with State-of-the-art Methods}
We compared with other methods on standard action localization datasets: UCF-101, J-HMDB21, and UCF-Sports.

\subsubsection{Comparison on the J-HMDB21 Dataset}
\label{sec:j-hmdb}
The comparison with other methods on the J-HMDB21 dataset is presented in Table~\ref{tab:compare}. 
The mAP is averaged over all three splits of the J-HMDB-21 dataset.
Our TPN model without TUN already consistently outperforms other comparable methods, regardless of the overlap threshold.
This comparison indicates that our proposed TPN can generate more accurate tubelet proposals than other methods.
Furthermore, utilizing a LSTM in the TUN, we achieve state-of-the-art results consistently over all thresholds. Some example action tubes are shown in Fig.~\ref{fig:visualization}. Note that we only visualize the action tubes which have action score higher than $0.2$ for visual clarity.

In the MR-TS R-CNN model~\cite{actiontubeseccv}, each of the two streams in the framework is expanded into a multi-region structure, therefore the  entire network needs to be retrained. However, in our case, since our TPN and LSTM are combined in a \textit{plug-in} style, the model trained for TPN can still be used, and the features are shared between the two parts to further reduce the computation. 

We also report action classification results in Table~\ref{tab:jhmdb_class}. We pick the action class with the highest action score among all the tubelets for each video as the video-level classification result.
Our proposed model achieves outperforms other tube-based methods on classification as well.

\subsubsection{Comparison on the UCF-Sports Dataset}

The localization results on the UCF-Sports dataset are reported in Table~\ref{tab:compare}.
Similar to the J-HMDB21 dataset, we consistently outperform other methods under different thresholds. Some example action tubes are shown in Fig.~\ref{fig:visualization_sports}. Note that again we only visualize the action tubes which have action score higher than $0.2$.

\subsubsection{ Comparison on the UCF-101 Dataset}
\label{sec:UCF-101}

The UCF-101 dataset is an untrimmed dataset. The action only occurs in some frames in each video. Therefore, it is also important to perform temporal localization.
However, recent works \cite{actiontubesbmvc,actiontubeseccv} either use simple heuristics or a sliding window approach to facilitate temporal localization.   A full treatment of temporal localization would be an interesting extension to our work.  To illustrate the potential of our method, we perform action localization over the entire untrimmed video. 
In this way, the precision and recall of the action tubes will be penalized because there are no ground-truth bounding boxes in some frames.
Only \cite{actiontubeseccv} reports the mAP on UCF-101 without temporal localization, therefore, we only compare to \cite{actiontubeseccv} on this dataset.
With temporal localization, the mAP of our proposed method could be further improved.
Some example ($>0.2$ score) action tubes are shown in Fig.~\ref{fig:visualization_101}.

In summary, the experiments show that our generic tubelet proposal paradigm consistently outperforms the class-specific action tube generation paradigm.
Compared with previous methods~\cite{actiontubeseccv,actiontubesbmvc}, our framework does not include any class-dependent hyper parameters, and is compact and easy to generalize. 

\section{Conclusion}
\label{sec:concl}

In this paper, a novel generic class-independent Tube Proposal Network is proposed.  This network can generate a set of generic class-independent tubelet proposals to localize the action regions in the video sequence precisely.
These generic tubelet proposals can be fed into various temporal understanding networks, to perform more sophisticated temporal understanding. 
The proposed framework is utilized in action localization tasks, which requires the system to both localize the action region precisely, and classify the action class correctly.
Our method consistently outperforms other methods across all three datasets. In our experiments, we observe that a limitation lies in handling the cases where multiple action instances are close to each other. Also, how to utilize the TPN as a backbone structure to facilitate temporal localization in untrimmed videos is an interesting direction. 
Further, these generic tubelets could be used in more complicated tasks such as interpreting scenes with multiple, interacting humans.  These directions will be addressed in future work. 

\bibliographystyle{ieee}
\bibliography{egbib}

\end{document}